\documentclass[11pt]{article}
\usepackage{amsmath}
\usepackage{graphicx}
\begin{document}
\title{A Computationally-Aware Multi-Objective Framework for Camera–LiDAR Calibration}
\author{Venkat Sai Raxit Karramreddy \\
Department of Electrical and Computer Engineering\\
Michagan State University\\
East Lansing, MI 48864
\and
Rangarajan Ramanujam \\
Independent Researcher \\
Irvine,CA 92618}
\date{}
\maketitle
\begin{abstract}
Accurate extrinsic calibration between LiDAR and camera sensors is important for reliable perception in autonomous systems. In this paper, we present a novel multi-objective optimization framework that jointly minimizes the geometric alignment error and computational cost associated with camera-LiDAR calibration. We optimize two objectives: (1) error between projected LiDAR points and ground-truth image edges, and (2) a composite metric for computational cost reflecting runtime and resource usage. Using the NSGA-II \cite{deb2002nsga2} evolutionary algorithm, we explore the parameter space defined by 6-DoF transformations and point sampling rates, yielding a well-characterized Pareto frontier that exposes trade-offs between calibration fidelity and resource efficiency. Evaluations are conducted on the KITTI dataset using its ground-truth extrinsic parameters for validation, with results verified through both multi-objective and constrained single-objective baselines. Compared to existing gradient-based and learned calibration methods, our approach demonstrates interpretable, tunable performance with lower deployment overhead. Pareto-optimal configurations are further analyzed for parameter sensitivity and innovation insights. A preference-based decision-making strategy selects solutions from the Pareto knee region to suit the constraints of the embedded system. The robustness of calibration is tested across variable edge-intensity weighting schemes, highlighting optimal balance points. Although real-time deployment on embedded platforms is deferred to future work, this framework establishes a scalable and transparent method for calibration under realistic misalignment and resource-limited conditions, critical for long-term autonomy, particularly in SAE L3+ vehicles receiving OTA updates.
\end{abstract}

\begin{center}
\end{center}

\section{Introduction}

The integration of LiDAR (Light Detection and Ranging) and camera data has become foundational in modern autonomous systems, enabling a robust understanding of the environment through multimodal sensing. This sensor fusion plays a vital role across a wide range of applications—autonomous driving, 3D reconstruction, urban planning, and environmental monitoring—by leveraging the complementary strengths of each modality. While LiDAR provides precise 3D geometric information, camera systems contribute high-resolution color and texture data. Together, these sensor systems offer more accurate object detection, enhanced depth perception, and detailed scene reconstruction, essential for navigation and decision-making in complex environments.

A core requirement for effective fusion is precise \textit{extrinsic calibration}—the spatial alignment of the LiDAR and camera coordinate frames through six parameters: three rotations $(\theta_{\text{yaw}}, \theta_{\text{pitch}}, \theta_{\text{roll}})$ and three translations $(x, y, z)$. Even slight miscalibration can propagate significant errors through perception pipelines, adversely affecting object detection, tracking, and map consistency. Sensor fusion using LiDAR and camera is foundational for autonomous systems. While LiDAR provides 3D structure, cameras offer rich visual context. Accurate extrinsic calibration between them is critical for robust perception

While traditional calibration approaches utilize controlled settings and predefined targets (e.g., checkerboards) \cite{zhang2004extrinsic}, such methods are impractical in deployed autonomous vehicles. These vehicles are subject to hardware shifts due to vibrations, thermal changes, or minor collisions, all of which can alter sensor alignment over time. In real-world production settings, particularly in SAE Level 3 and higher vehicles, these sensors are permanently mounted, and the opportunity for manual recalibration is extremely limited after deployment.

In addition to this, due to the nature of embedded processors in production vehicles, which typically have limited computational resources. As vehicles increasingly receive over-the-air software updates with new functionalities and safety features, resource availability becomes even more constrained. Yet, calibration quality must remain high, especially because safety-critical features such as emergency braking, lane keeping, and adaptive cruise control depend directly on accurate sensor alignment. The impact of the TBD allowable error, however low it may be, can be extended to incorrect actuation or unsafe scenarios in the overall vehicle trajectory and control. Corresponding Hazard analysis can be carried out and metrics developed to provide a nominal target for the camera systems to tune towards and if the system goes out of sync, this can be fixed by preemptively bringing the vehicle to the dealership/OEM to get it fixed before any big mishap/accident. 

Calibration errors must often be addressed while the vehicle is in motion, such as during long drives or dynamic scenari 

To address these constraints, we propose a multi-objective optimization framework that balances calibration accuracy (measured by Chamfer error) \cite{wu2021densityawarechamferdistancecomprehensive} and computational cost. We employ the NSGA-II algorithm \cite{deb2002nsga2} to explore trade-offs between these competing objectives, identify Pareto-optimal solutions, and gain insights into calibration behavior through an innovization study. This approach enables robust, data-driven calibration adaptable to resource-constrained platforms and dynamic operational contexts.

Our key contributions include:
\begin{itemize}
    \item A practical multi-objective formulation of the camera–LiDAR calibration problem that considers both geometric alignment and real-time processing constraints.
    \item Application of NSGA-II to compute Pareto-optimal extrinsic parameters across various initializations.
    \item An innovization-based analysis to uncover parameter patterns and design insights from optimal solutions.
    \item A weighted-base loss function calculations that further optimizes the error.
\end{itemize}

\subsection{Coordinate Transformation and Depth Map Generation}

The projection of 3D world points onto 2D image space requires the use of multiple coordinate systems. Typically, a point originating in the world coordinate frame undergoes a sequence of transformations: first to the camera coordinate frame, then to the normalized image plane, and finally to pixel coordinates. This chain of transformations ensures that physical 3D data can be properly mapped and interpreted in 2D images.

The conversion from the world coordinate frame $(x_w, y_w, z_w)$ to the camera coordinate frame $(x_c, y_c, z_c)$ is governed by both rotation and translation\cite{hartley2003multiple}. A combined transformation matrix is used to encode these effects, where the rotation matrix $R$ accounts for orientation changes (e.g., yaw, pitch, and roll), and the translation vector $T$ represents the spatial shift between coordinate frames:

\[
\begin{bmatrix}
x_c \\
y_c \\
z_c \\
1
\end{bmatrix}
=
\begin{bmatrix}
R & T \\
0 & 1
\end{bmatrix}
\begin{bmatrix}
x_w \\
y_w \\
z_w \\
1
\end{bmatrix}
\]

Following this, the 3D coordinates are projected onto the 2D image plane using the perspective projection model:

\[
x = \frac{x_c}{z_c}, \quad y = \frac{y_c}{z_c}
\]

To transition from the image plane to the pixel coordinate system, the intrinsic parameters of the camera—such as focal lengths $(f_x, f_y)$ and the optical center $(u_0, v_0)$—are used via the intrinsic matrix $K$:

\[
\begin{bmatrix}
u \\
v \\
1
\end{bmatrix}
=
\begin{bmatrix}
f_x & 0 & u_0 \\
0 & f_y & v_0 \\
0 & 0 & 1
\end{bmatrix}
\begin{bmatrix}
x \\
y \\
1
\end{bmatrix}
\]

Putting this all together, the transformation from a world coordinate point to a pixel location in the image can be written as:

\[
z_c \begin{bmatrix}
u \\
v \\
1
\end{bmatrix}
=
K_1 \cdot K_2 \cdot
\begin{bmatrix}
x_w \\
y_w \\
z_w \\
1
\end{bmatrix}
\]

Where $K_2$ is the extrinsic matrix $\begin{bmatrix} R & T \\ 0 & 1 \end{bmatrix}$ and $K_1$ is the camera intrinsic matrix extended to 3x4 form.

\subsubsection*{Depth Map Generation}

To derive a depth map from a 3D LiDAR point cloud, each point $(X, Y, Z)$ is first transformed into the camera frame using the extrinsic parameters \cite{schultz2020robotics}. This transformation is given by:

\[
\begin{bmatrix}
X_c \\
Y_c \\
Z_c
\end{bmatrix}
=
R \cdot
\begin{bmatrix}
X \\
Y \\
Z
\end{bmatrix}
+ T
\]

Once in the camera frame, the points are projected into 2D image space using the camera intrinsics:

\[
\begin{bmatrix}
x' \\
y' \\
z'
\end{bmatrix}
=
K
\begin{bmatrix}
X_c \\
Y_c \\
Z_c
\end{bmatrix}
\quad \text{with} \quad
K =
\begin{bmatrix}
f_x & 0 & u_0 \\
0 & f_y & v_0 \\
0 & 0 & 1
\end{bmatrix}
\]

The final pixel coordinates are obtained by normalizing:

\[
u = \frac{x'}{z'} , \quad v = \frac{y'}{z'}
\]

\begin{figure}[h]
    \centering
    \includegraphics[width=0.7\textwidth]{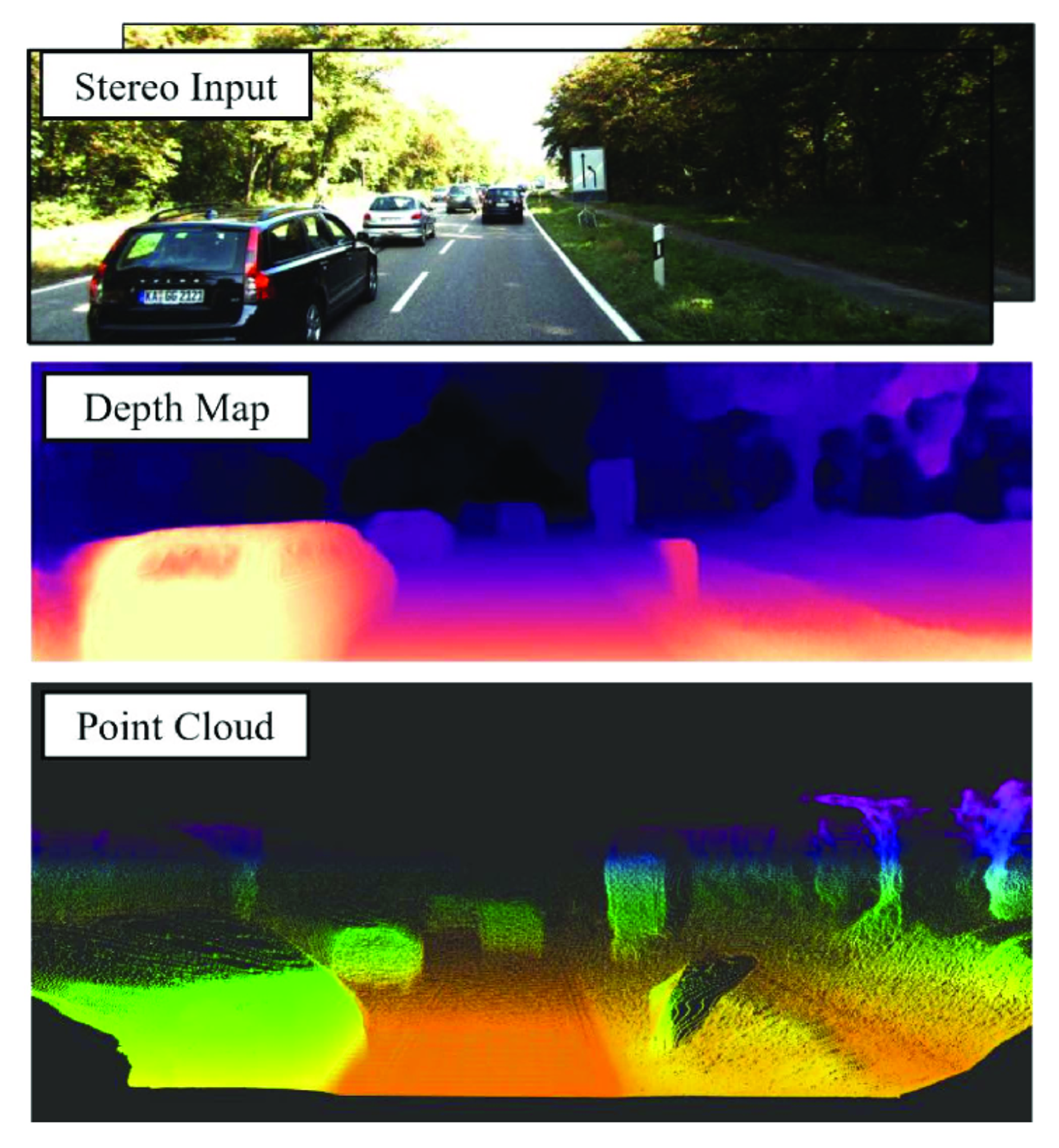}
    \caption{Illustration of data transformation from stereo image input to depth map and projected LiDAR point cloud.}
    \label{fig:depthfusion}
\end{figure}

The depth at each pixel location $(u, v)$ is typically represented by $Z_c$, corresponding to the distance of the point along the camera’s optical axis. This transformation enables sparse LiDAR data to be structured into a dense 2D format aligned with the camera’s viewpoint \cite{karramreddy2024validationexplorationmultimodal}, as illustrated in Figure~\ref{fig:depthfusion}.

\section{Problem Formulation}

\begin{figure}
    \centering
    \includegraphics[width=0.9\linewidth]{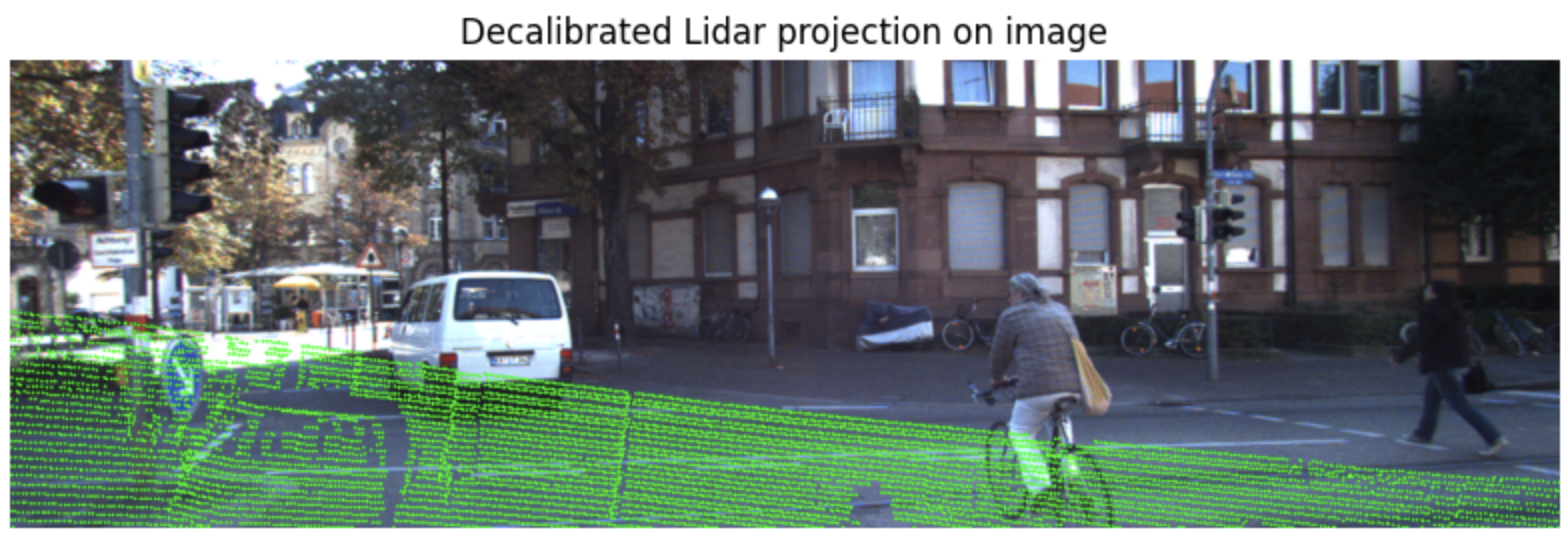}
    \caption{Projection of LiDAR points onto the image plane using an initially de-calibrated transformation. Significant misalignment can be observed between the LiDAR points and the image features.}
    \label{fig:initial_projection}
\end{figure}

After projecting initially de-calibrated LiDAR points onto the image plane \cite{zhou2018open3d}, significant misalignment is typically observed, as shown in Figure~\ref{fig:initial_projection}. The points appear offset from image features, making sensor fusion completely useless. This misalignment occurs due to errors in the extrinsic calibration between LiDAR and camera. To address this, we formulate a multi-objective optimization problem that aims to refine the extrinsic transformation parameters and restore alignment accuracy.

The calibration process is guided by two competing objectives: improving the geometric alignment between sensor modalities and reducing the computational cost involved in evaluating calibration quality. These objectives inherently conflict, as higher precision often requires processing a larger volume of data, which increases runtime and memory usage.

\subsection{Objective Functions}

The optimization considers the following two goals:

\begin{itemize}
    \item \textbf{Calibration Accuracy:}  
    The first objective is to minimize the alignment error between LiDAR-projected features and ground-truth image features. This is achieved by quantifying the discrepancy between edge features derived from the LiDAR point cloud and those extracted from the camera image \cite{besl1992icp} \cite{an2020geometric}. Improved alignment corresponds to better calibration fidelity.

    \item \textbf{Computational Resources:}  
    The second objective is to reduce the computational load associated with evaluating candidate calibrations. This includes the time required for projecting LiDAR points, calculating distances or errors, and any memory used in storing intermediate computations. A solution that uses fewer resources is preferable, especially for real-time or embedded applications.
\end{itemize}

\subsection{Design Variables}

The optimization is carried out over a set of seven design variables that define the transformation between the LiDAR and camera frames, along with a sampling parameter for efficiency:

\begin{itemize}
    \item \textbf{Roll} (rotation around the X-axis), constrained within $[-25^\circ, 25^\circ]$
    \item \textbf{Pitch} (rotation around the Y-axis), constrained within $[-25^\circ, 25^\circ]$
    \item \textbf{Yaw} (rotation around the Z-axis), constrained within $[-25^\circ, 25^\circ]$
    \item \textbf{X translation}, within $[-1.5\, \text{m}, 1.5\, \text{m}]$
    \item \textbf{Y translation}, within $[-1.5\, \text{m}, 1.5\, \text{m}]$
    \item \textbf{Z translation}, within $[-1.5\, \text{m}, 1.5\, \text{m}]$
    \item \textbf{Number of LiDAR points} used for evaluation, which influences both accuracy and resource cost
\end{itemize}

These variables define the solutions evaluated during the optimization. The addition of the number of LiDAR points as a decision variable allows the optimization to find a balance between detail in the error metric and computational efficiency.  

The constraints on these variables are purposefully large to simulate realistic worst-case scenarios. Miscalibrations can occur due to sensor drift, hardware reinstallation, or mechanical shock. By allowing large deviations in rotation and translation, we ensure that the optimization method is tested against significant initial misalignments and is capable of recovering accurate calibration even under adverse conditions.

\section{Optimal Design Problem}

Based on the problem discussed above, the optimization variables are defined as:

\[
\mathbf{X} = [x, y, z, \theta_{\text{yaw}}, \theta_{\text{pitch}}, \theta_{\text{roll}}, n]
\]
where:
\begin{itemize}
    \item $(x, y, z)$ denotes the translation vector from the LiDAR frame to the camera frame,
    \item $(\theta_{\text{yaw}}, \theta_{\text{pitch}}, \theta_{\text{roll}})$ represent the Euler angles defining the rotation,
    \item $n$ is the number of LiDAR points used during the iteration for projection and evaluation.
\end{itemize}

\subsection{Chamfer Distance Error}

The first objective measures the geometric misalignment between the projected LiDAR points and the ground-truth edge map in the camera image. The Chamfer distance is defined as:
\[
E_{\text{chamfer}}(\mathbf{X}) = \frac{1}{n} \sum_{p \in P_{\text{GT}}} \min_{q \in P_{\text{est}}(\mathbf{X})} \|p - q\|^2
\]
where $P_{\text{GT}}$ is the set of ground-truth edge points and $P_{\text{est}}(\mathbf{X})$ is the set of projected LiDAR points under the transformation defined by $\mathbf{X}$. This objective penalizes spatial misalignment and favors accurate edge correspondence.

\subsection{Computational Cost}

The second objective accounts for the computational burden of processing the selected LiDAR points. It is modeled as:
\[
E_{\text{comp}}(\mathbf{X}) = t_{\text{norm}}(\mathbf{X}) + m_{\text{norm}}(\mathbf{X})
\]
where:
\begin{itemize}
    \item $t_{\text{norm}}(\mathbf{X})$ is the normalized computation time (e.g., projection and matching),
    \item $m_{\text{norm}}(\mathbf{X})$ is the normalized memory or data usage.
\end{itemize}

Since both time and memory scale with $n$, this objective captures the trade-off between computational cost and potential gains in accuracy from using more LiDAR points.

\subsection{Multi-Objective Optimization Problem}

The complete multi-objective optimization problem is formulated as:
\[
\min_{\mathbf{X}} \left( E_{\text{chamfer}}(\mathbf{X}),\; E_{\text{comp}}(\mathbf{X}) \right)
\]
subject to the following bounds:
\[
x, y, z \in [-1.5\,\text{m}, +1.5\,\text{m}], \quad
\theta_{\text{yaw}}, \theta_{\text{pitch}}, \theta_{\text{roll}} \in [-25^\circ, +25^\circ], \quad
n \in [n_{\text{min}}, n_{\text{max}}]
\]

With this formulation, we can evaluate calibration accuracy and computational cost simultaneously using NSGA-II, which is particularly effective for exploring trade-offs in Pareto-optimal solutions.

All optimization variables are continuous and bounded. \texttt{SBX crossover}, \texttt{Polynomial mutation}, and \texttt{float sampling} are used as GA operators. Infeasible or out-of-bound solutions are prevented by using \texttt{FloatRandomSampling} and the \texttt{eliminate\_duplicates} option. No additional constraints beyond the variable bounds were imposed. The implementation was carried out in Python using the \texttt{pymoo} library.

\section{Simulation Results}
\subsection{Optimization Parameters and Setup}

The optimization was performed using the NSGA-II algorithm implemented through the PyMOO framework \cite{pymoo}. The data that was used for this evaluation is imported from raw dataset sequences from KITTI dataset \cite{Geiger2013IJRR}. Key parameters used in the simulation include:

\begin{itemize}
    \item \textbf{Population size:} 1000 (increased from standard values to obtain a more diverse Pareto front)
    \item \textbf{Number of generations:} 200
    \item \textbf{Crossover operator:} Simulated Binary Crossover (SBX) with probability 0.9 and distribution index 15
    \item \textbf{Mutation operator:} Polynomial Mutation (PM) with distribution index 20
    \item \textbf{Termination criteria:} Fixed number of generations (200)
    \item \textbf{Variable bounds:} Rotation range of $\pm$0.1 radians for roll, pitch, and yaw
    \item \textbf{Computational complexity factor:} 3 iterations per evaluation
\end{itemize}

The optimization problem was formulated with two objectives: (1) minimizing the Chamfer distance between projected point cloud edges and ground truth edges to improve calibration accuracy, and (2) minimizing computational resource usage, defined as a weighted combination of execution time and memory consumption.

A total of \textbf{200,000 function evaluations} were performed during the optimization, with each evaluation assessing both objectives. The algorithm maintained a non-dominated solution set throughout, forming the final Pareto front.

\subsection{Implementation Challenges and Solutions}

Several challenges were encountered during implementation:

\begin{itemize}
    \item \textbf{Computational Complexity Management:} Edge detection and point cloud projection were computationally intensive. A resource measurement framework using the \texttt{psutil} library was implemented to accurately track execution time and memory usage.
    
    \item \textbf{Data Management:} Efficient tracking of metrics over 200,000 evaluations required robust data structures. Custom callback functions were implemented to log statistics at each generation.
    
    \item \textbf{Instability in Chamfer Distance Evaluation:} Brute-force nearest-neighbor search led to numerical instability. This was resolved by switching to \texttt{scipy.spatial.cKDTree} for efficient and stable distance computation.
\end{itemize}

The outcomes of the optimization process are visualized using two key plots: the Pareto front of non-dominated solutions, and the distribution of solutions in the 3D parameter space.

\subsection*{Pareto Front Visualization}

\begin{figure}
    \centering
    \includegraphics[width=0.8\textwidth]{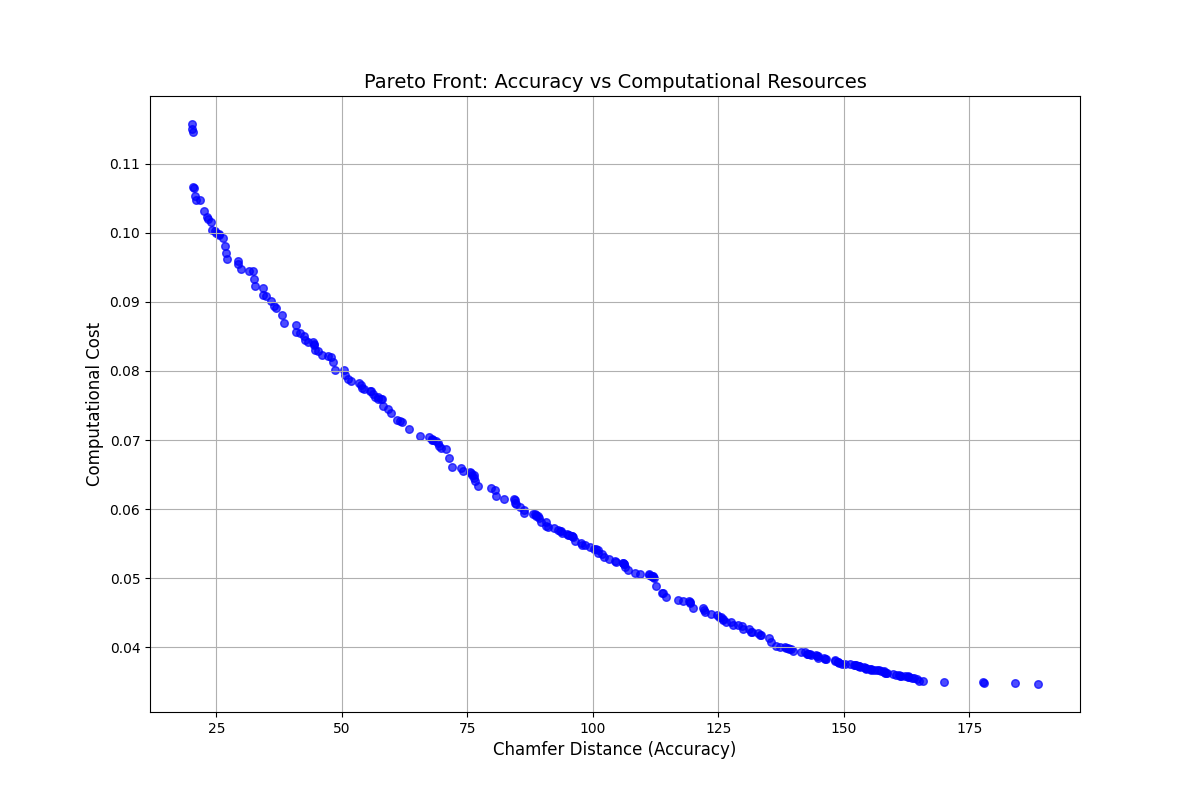}
    \caption{Pareto front showing the trade-off between Chamfer error and computational cost.}
    \label{fig:pareto_front}
\end{figure}

Figure~\ref{fig:pareto_front} illustrates the Pareto front obtained after running the NSGA-II algorithm across 200 generations. Each point represents a non-dominated solution in the objective space, where the x-axis corresponds to the Chamfer error and the y-axis represents the normalized computational cost. The trade-off between the two objectives is clearly visible: solutions with lower Chamfer error generally exhibit higher computational cost, while faster solutions often compromise on calibration accuracy. The smooth spread of solutions along the front indicates the algorithm's ability to maintain diversity and convergence throughout the optimization process.

\subsection*{3D Solution Space}

\begin{figure}
    \centering
    \includegraphics[width=0.8\textwidth]{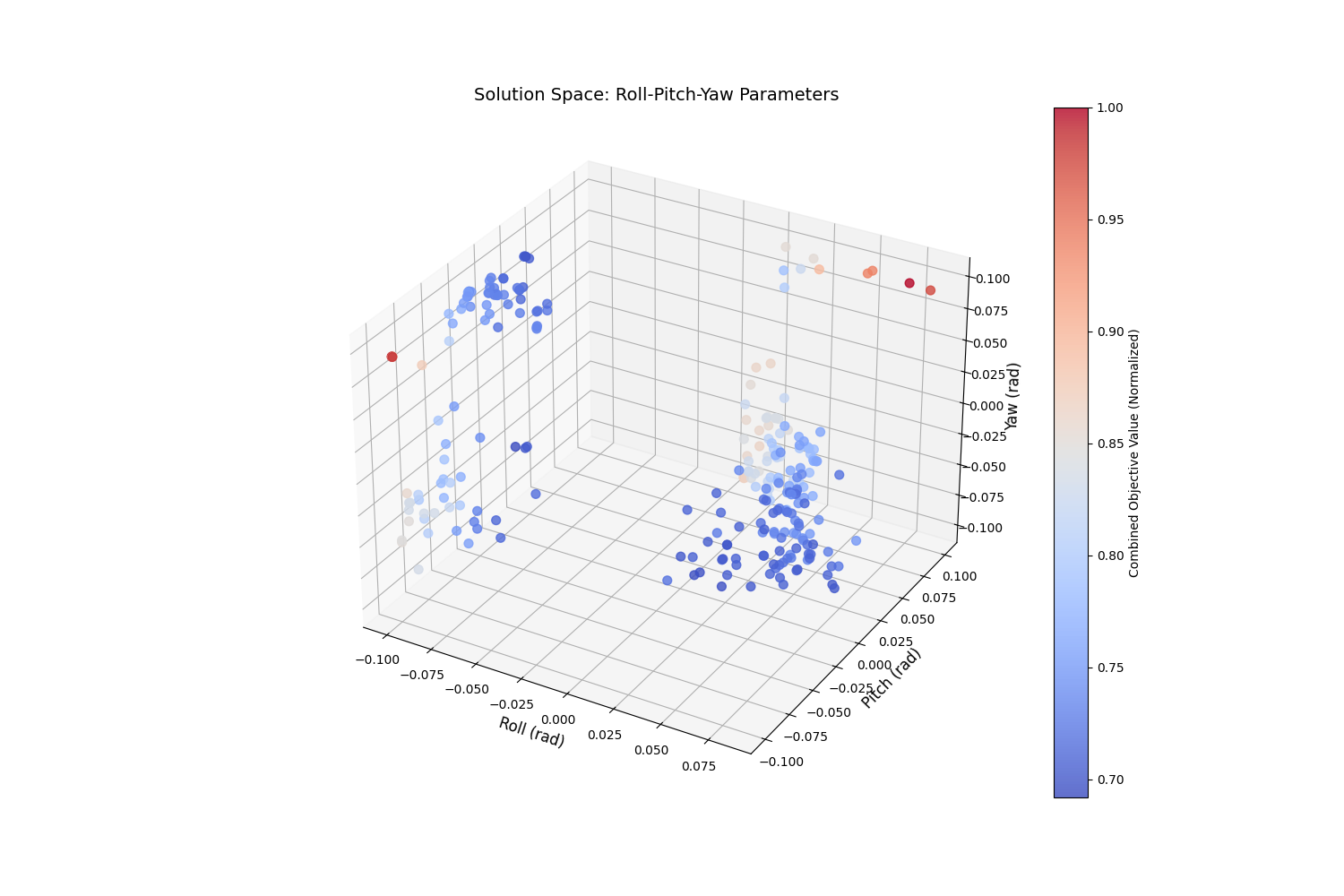}
    \caption{3D solution space showing the distribution of extrinsic parameters across Pareto-optimal configurations.}
    \label{fig:solution_space_3d}
\end{figure}

To better understand the distribution of extrinsic calibration parameters across optimal solutions, a 3D visualization of the solution space is provided in Figure~\ref{fig:solution_space_3d}. This figure plots the optimized variables (e.g., translation and rotation values) across the final Pareto-optimal set. It provides insight into the structure and clustering behavior of high-performing solutions in the decision space. Such visualization is especially useful for identifying regions of parameter space that yield balanced performance with respect to both objectives.

\subsection{Verification by Single-objective Optimization Methods}
\begin{figure}
    \centering
    \includegraphics[width=0.9\textwidth]{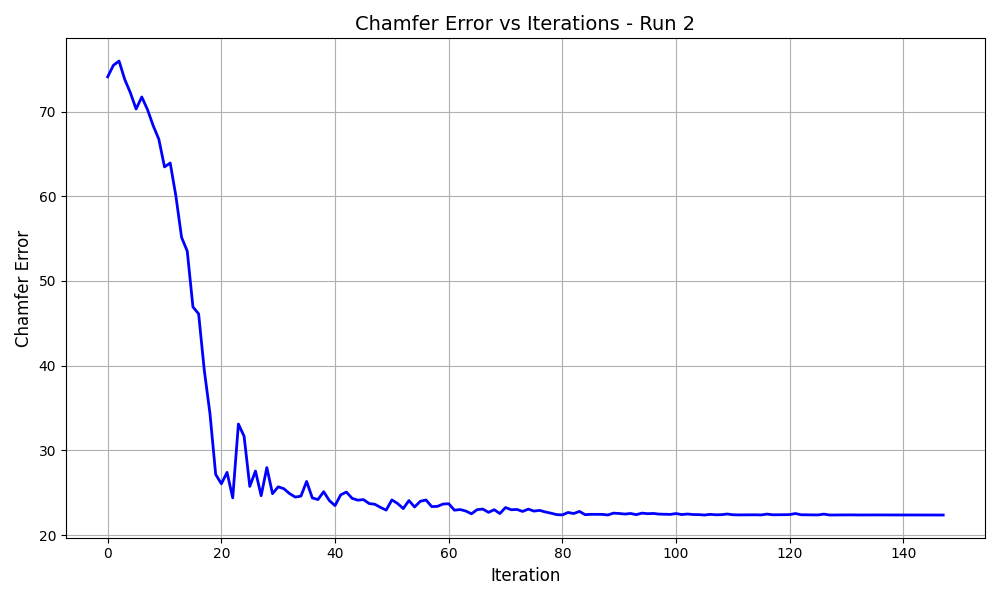}
    \caption{Single-objective optimization of Chamfer distance over iterations. The plot demonstrates convergence behavior when optimizing only for Chamfer error, without considering computational cost.}
    \label{fig:single_obj_chamfer}
\end{figure}

As shown in Figure~\ref{fig:single_obj_chamfer} the error decreases for iterations when we perform a single objective optimization.

\begin{figure}
    \centering
    \includegraphics[width=0.9\textwidth]{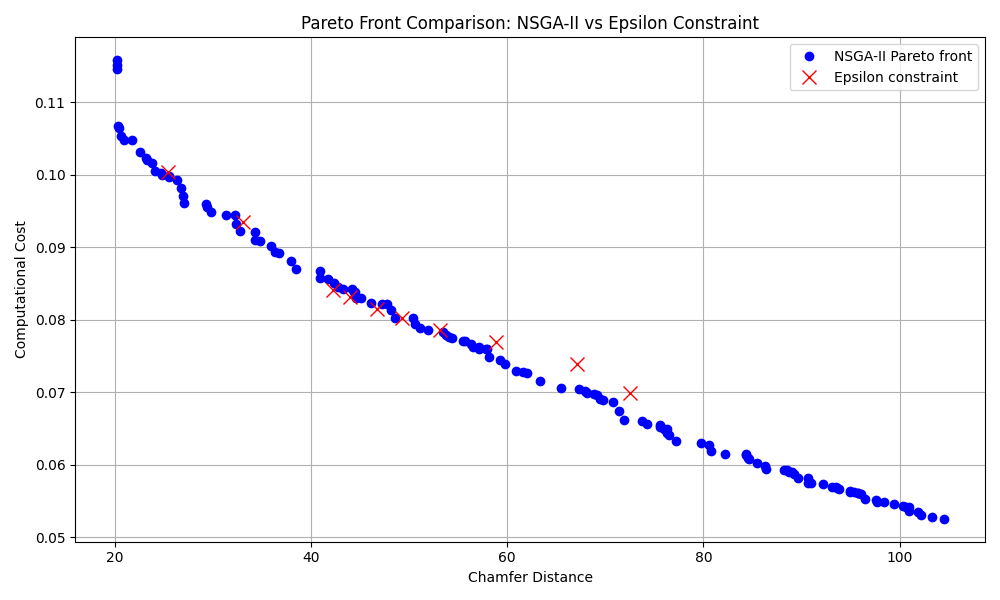}
    \caption{Comparison of NSGA-II Pareto front and Epsilon-Constraint solutions. Blue dots represent Pareto-optimal solutions obtained from NSGA-II, while red crosses represent solutions selected by fixing computational cost and minimizing Chamfer distance.}
    \label{fig:epsilon_constraint}
\end{figure}

The red crosses in the Figure~\ref{fig:epsilon_constraint} represent the feasible solutions obtained by applying epsilon constraints on computational cost. We are constraining the computational resources at various points and minimizing the chamfer error at those points. The figure illustrates the points closely follow the Pareto front that was found through NSGA-II. The slight deviation of epsilon points at lower level of computational resource values is due to the inability to reach a steady state value due to large deviation in the initial decalibration. Moreover, some of the deviation can also be attributed to the number of iterations that was used to optimize. Lower iterations were used in epsilon constraint method when compared to the NSGA-II method.

\subsection{Preference-based Decision-making}
\begin{figure}
    \centering
    \includegraphics[width=0.9\textwidth]{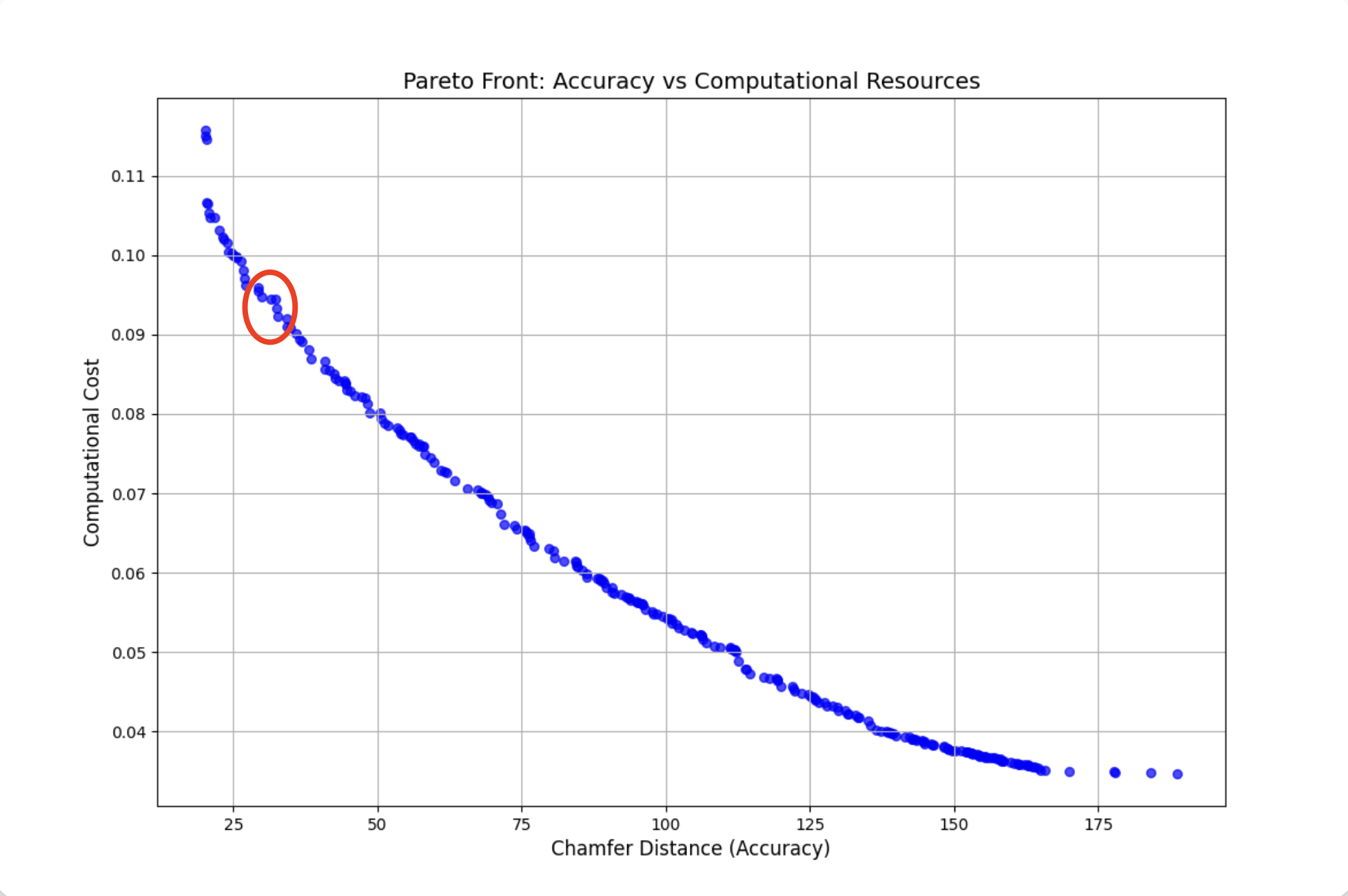}
    \caption{Decision-making based on the Pareto front.}
    \label{fig:knee_point_decision}
\end{figure}

To determine the most suitable calibration configuration for deployment, we rely on proper Pareto optimality principles alongside application-specific constraints. Our goal is to ensure that the selected solution maintains a low Chamfer error, providing a calibration result that closely matches the ground truth. The final selection is made from a knee region on the Pareto front, which represents a zone of balanced trade-off. Here, small sacrifices in error yield disproportionately large gains in computational cost, making this region an ideal compromise. Figure~\ref{fig:knee_point_decision} highlights this area, which satisfies both the accuracy threshold and efficiency requirements, making it optimal for practical deployment scenarios.

\begin{figure}
    \centering
    \includegraphics[width=0.9\textwidth]{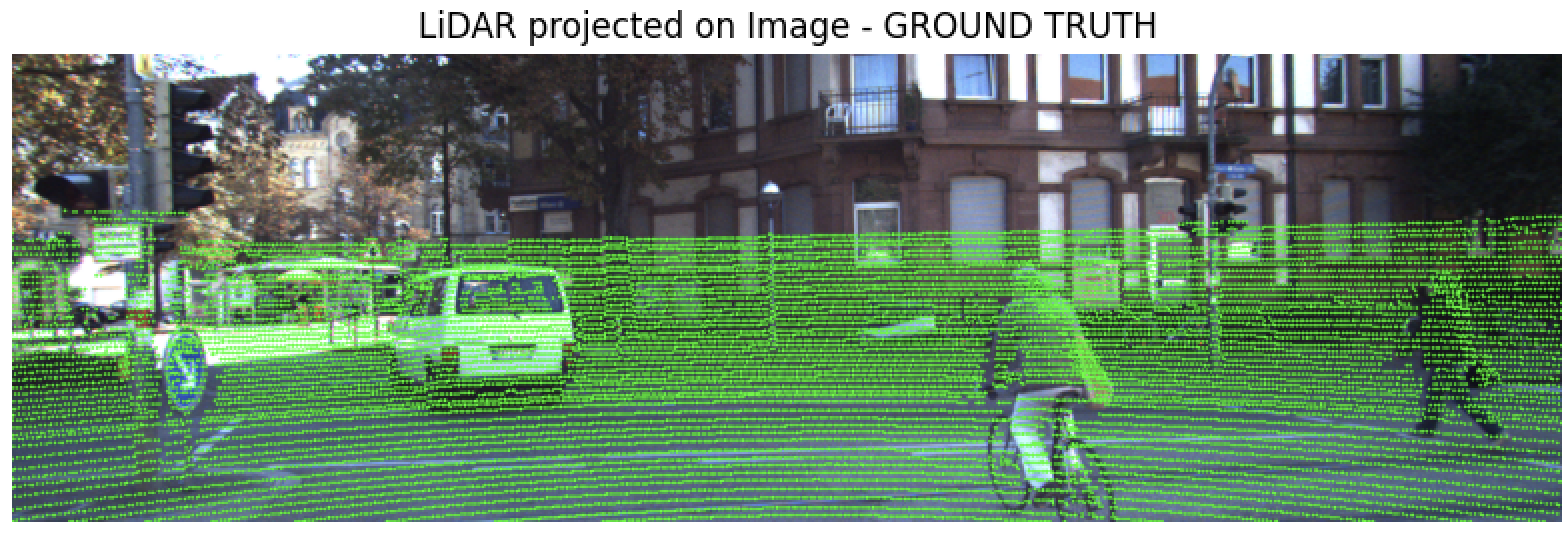}
    \caption{Original ground truth projection.}
    \label{fig:Original ground truth projection}
\end{figure}

\begin{figure}
    \centering
    \includegraphics[width=0.9\textwidth]{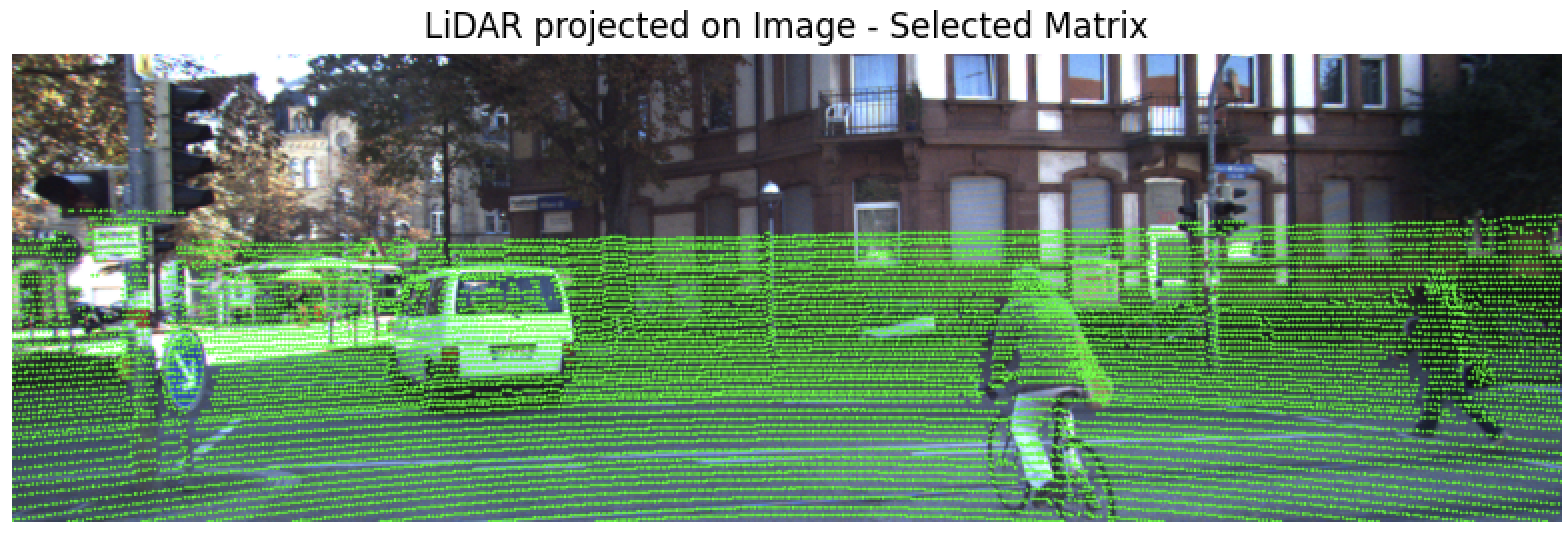}
    \caption{Projection of LiDAR point cloud based on selected pareto front solution.}
    \label{fig:projection of our solution matrix}
\end{figure}

Figure~\ref{fig:Original ground truth projection} shows the projection of the ground truth of the LiDAR point cloud onto the image plane. Figure~\ref{fig:projection of our solution matrix} shows the projection of the LiDAR point cloud onto the image plane after using the calibration matrix from the selected Pareto front.

\subsection{``Innovization'' Study}
In our case, we examined the relationship between the calibration parameters—namely, the rotation angles ($\theta_{\text{yaw}}, \theta_{\text{pitch}}, \theta_{\text{roll}}$) and position values ($x, y, z$)—and their influence on the Chamfer distance error. The analysis of the Pareto-optimal solutions showed that no dominant interaction was observed among the variables. Each parameter independently contributed to the error metric.

Moreover, all three rotation variables exhibited similar change on the error magnitude, and it is same for the three position variables. This indicates that no individual transformation dimension disproportionately drives the calibration performance, and changes in each variable affect the Chamfer distance in comparable ways. The innovization study did not show any correlation among the decision variables or between the variables and the objectives in this problem.

\subsection{Robustness Study}
\begin{figure}[h]
    \centering
    \includegraphics[width=0.7\textwidth]{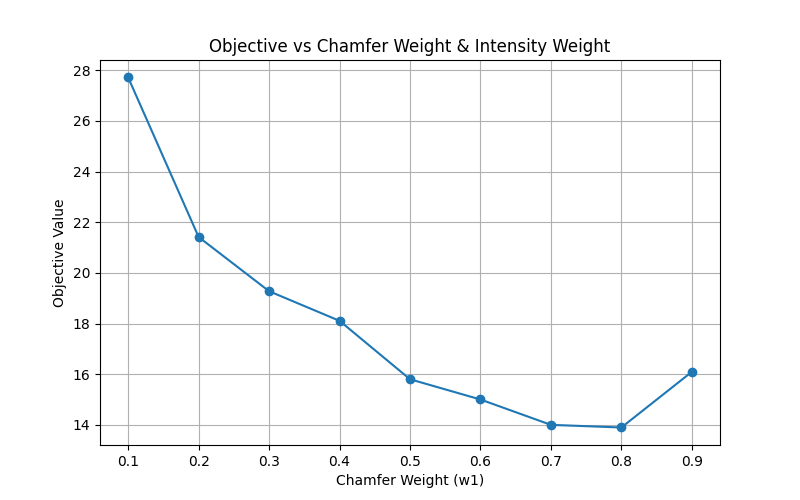}
    \caption{Objective value vs. weights}
    \label{fig:robustness_plot}
\end{figure}

A robustness analysis is conducted to evaluate the sensitivity of the optimization results to different combinations of weights applied to two Chamfer distance metrics: one derived from edge maps and the other from intensity maps. The overall objective function was defined as a weighted sum of the two metrics:  
\[
\text{Error} = w_1 \cdot E_{\text{chamfer\_edge}} + w_2 \cdot E_{\text{chamfer\_intensity}} \quad \text{with} \quad w_1 + w_2 = 1.
\]
To assess the effect of the weights, the optimization was repeated for multiple values of \( w_1 \) ranging from 0.1 to 0.9 and \( w_2 \) ranging from 0.9 to 0.1. The corresponding error values were recorded and analyzed. We observed that when edge-based Chamfer distance was assigned a low weight (e.g., \( w_1 = 0.1, w_2 = 0.9 \)), the resulting objective value was significantly higher (27.71), indicating poor calibration performance. As the weight on edge-based information increased, the error consistently decreased to 13.997 at \( w_1 = 0.7, w_2 = 0.3 \), and finally reaching a minimum of 13.893 at \( w_1 = 0.8, w_2 = 0.2 \), after which it began to increase again at \( w_1 = 0.9, w_2 = 0.1 \) (error = 16.03). This U-shaped trend in the error curve reveals that neither metric alone is sufficient to achieve optimal calibration. Instead, the results highlight the need for a balanced approach, where both geometric (edge-based) and photometric (intensity-based) \cite{9341147} information are leveraged. The best performance was achieved when edge information was given a dominant but not exclusive influence in the objective function (i.e., 80\% edge and 20\% intensity). This implies that while edge structures offer stronger spatial alignment cues, incorporating intensity data enhances calibration robustness by capturing subtle photometric consistencies. The robustness study confirms that weighting parameters significantly influence the final solution and that an intermediate weighting strategy yields the most reliable and accurate calibration results. The variation in objective values as a function of edge weight is shown in Figure~\ref{fig:robustness_plot}.

\subsection{Safety-Critical Impact in Automotive Applications}

In production-grade autonomous and semi-autonomous vehicles, extrinsic calibration between camera and LiDAR sensors directly influences safety-critical perception tasks such as object detection, collision avoidance, lane keeping, and adaptive cruise control. Even minor misalignment between sensor modalities can cascade into spatial errors in 3D object localization, leading to degraded decision-making or unsafe maneuvers.

From a safety systems engineering perspective, our proposed computationally aware calibration framework contributes to:

\begin{itemize}
    \item \textbf{Perception Reliability in Safety Chains:} \\
    Accurate projection of LiDAR points onto camera frames ensures precise sensor fusion, reducing false negatives/positives in object detection systems. This leads to more stable bounding boxes and scene reconstructions, minimizing ambiguity in downstream planning modules.

    \item \textbf{Fail-Operational Behavior:} \\
    By incorporating computational cost as an optimization objective, our framework ensures that calibration routines operate within bounded CPU/GPU budgets, supporting redundancy controllers without jeopardizing real-time execution of critical functions (e.g., braking, steering).

    \item \textbf{Safety Standards Compliance:} \\
    The deterministic and interpretable nature of our NSGA-II-based optimization aligns well with traceability and verification requirements defined in ISO 26262 (Functional Safety) and ISO/PAS 21448 (SOTIF). Future versions may integrate ASIL-aware calibration thresholds or fault detection routines to support graded safety strategies.

    \item \textbf{OTA Resilience and Post-Event Recovery:} \\
    As modern vehicles rely on over-the-air (OTA) software updates, our modular framework supports autonomous recalibration following updates or sensor perturbations. This promotes self-healing calibration and enhances uptime without manual servicing.
\end{itemize}

\noindent
\textbf{Real-World Example:} Consider a Level 3 autonomous highway driving system where camera–LiDAR alignment is off by even a few degrees. This misalignment may lead to incorrect distance estimation or lane association, adversely affecting safety decisions such as lane changes or emergency braking. A lightweight background recalibration mechanism is therefore not only beneficial for perception fidelity but also essential for safe and certifiable autonomy.

\section{Future Extensions}

Beyond improving the robustness of edge-intensity weighting in loss functions, future work will integrate metrics such as SSIM(perceptual similarity), photometric consistency(illumination invariance), smoothed L1 loss(robust regression) \cite{lv2021lccnetlidarcameraselfcalibration} and PSO-based losses such as Point-PSO and Pose-PSO (feature and pose alignment) \cite{9802778}. 

To combine these metrics into a unified objective, we propose a weighted multi-loss framework:

\[
\mathcal{L}_{\text{total}} = \sum_i w_i \cdot L_i(x, y, z, \theta_{\text{yaw}}, \theta_{\text{pitch}}, \theta_{\text{roll}})
\]

This formulation allows dynamic balancing of different error weights, suitable for different scenarios. Weighting strategies may also include fixed heuristics, uncertainty-based weighting, or Pareto-based approaches.

In addition, future work can potentially focus on developing objective metrics related to how calibration accuracy affects downstream ADAS tasks like object detection and lane tracking and integrating safety mechanisms aligned with ISO 26262 and SOTIF - including ASIL-aware thresholds, misalignment fallback strategies, and OTA-enabled self-healing recalibration. The weighted objective function offers a realistic way of integrating safety into the design.

\section{Conclusions}

In this paper, we addressed the problem of extrinsic calibration between LiDAR and camera systems using a multi-objective optimization framework. By minimizing both the Chamfer alignment error and computational cost, we explored a set of Pareto-optimal solutions that reflect the trade-offs between accuracy and efficiency. The optimization was performed over the six degrees of freedom of the sensor transformation, and the resulting Pareto front provided actionable insights into optimal calibration choices under resource constraints.

Using NSGA-II, the framework generates interpretable Pareto-optimal solutions over 6-DoF transformations and LiDAR sampling rates. Evaluations on the KITTI dataset validate its effectiveness against significant initial misalignments, with results outperforming traditional and learning-based calibration methods.
Crucially, the framework supports safety-critical functionalities such as accurate object detection and lane estimation, which depend heavily on sensor alignment. Its lightweight nature ensures it can operate within the resource limits of embedded systems, supporting fail-operational behavior, OTA resilience, and alignment with ISO 26262 and SOTIF standards.

This method lays the foundation for robust, scalable calibration in real-world autonomous systems. Future extensions will incorporate additional loss functions, such as SSIM and photometric error, and adopt a weighted loss formulation to further improve resilience and adaptability across diverse environments and sensor conditions. Future developments will continue to integrate safety-oriented thresholds, perception impact validation, and real-time hardware deployment, making this a foundational enabler of certifiable, autonomous vehicle safety.

\end{document}